\documentclass[10pt,twocolumn,letterpaper]{article}

\usepackage{cvpr}              % To produce the CAMERA-READY version
\usepackage{graphicx}
\usepackage{amsmath}
\usepackage{amssymb}
\usepackage{booktabs}
\usepackage{multirow}
\usepackage{ulem}
\usepackage[justification=centering]{caption}

\usepackage[accsupp]{axessibility}  % Improves PDF readability for those with disabilities.

\usepackage[pagebackref,breaklinks,colorlinks]{hyperref}

\usepackage[capitalize]{cleveref}
\crefname{section}{Sec.}{Secs.}
\Crefname{section}{Section}{Sections}
\Crefname{table}{Table}{Tables}
\crefname{table}{Tab.}{Tabs.}

\def\cvprPaperID{112} % *** Enter the CVPR Paper ID here
\def\confName{CVPR}
\def\confYear{2023}

\begin{document}

\title{Breaking Through the Haze: An Advanced Non-Homogeneous Dehazing Method based on Fast Fourier Convolution and ConvNeXt}

\author{Han Zhou$^1$, Wei Dong$^2$, Yangyi Liu$^1$, Jun Chen$^1$\\
$^1$Department of Electrical and Computer Engineering, Mcmaster University, Hamilton, Canada\\
$^2$Department of Electrical and Computer Engineering, University of Alberta, Edmonton, Canada\\
{\tt\small zhouh115@mcmaster.ca}, {\tt\small wdong1745376@gmail.com}, {\tt\small \{liu5, chenjun\}@mcmaster.ca}
%\and
}
\maketitle

%%%%%%%%% ABSTRACT
\begin{abstract}
   Haze usually leads to deteriorated images with low contrast, color shift and structural distortion. We observe that many deep learning based models exhibit exceptional performance on removing homogeneous haze, but they usually fail to address the challenge of non-homogeneous dehazing. Two main factors account for this situation. Firstly, due to the intricate and non uniform distribution of dense haze, the recovery of structural and chromatic features with high fidelity is challenging, particularly in regions with heavy haze. Secondly, the existing small scale datasets for non-homogeneous dehazing are inadequate to support reliable learning of feature mappings between hazy images and their corresponding haze-free counterparts by convolutional neural network (CNN)-based models. To tackle these two challenges, we propose a novel two branch network that leverages 2D discrete wavelete transform (DWT), fast Fourier convolution (FFC) residual block and a pretrained ConvNeXt model. Specifically, in the DWT-FFC frequency branch, our model exploits DWT to capture more high-frequency features. Moreover, by taking advantage of the large receptive field provided by FFC residual blocks, our model is able to effectively explore global contextual information and produce images with better perceptual quality. In the prior knowledge branch, an ImageNet pretrained ConvNeXt as opposed to Res2Net is adopted. This enables our model to learn more supplementary information and acquire a stronger generalization ability. The feasibility and effectiveness of the proposed method is demonstrated via extensive experiments and ablation studies. The code is available at \url{https://github.com/zhouh115/DWT-FFC}.
\end{abstract}

%%%%%%%%% BODY TEXT
\section{Introduction}
\label{sec:intro}
As a natural phenomenon, haze usually heavily reduces the visibility, resulting in blurred hazy images with low contrast, color shift and structural distortion. Various intelligent applications, like object detection \cite{object-detection} and autonomous driving \cite{auto-driving}, need to operate normally in hazy conditions, which necessitates the restoration of missing information from hazy images. As a consequence, image dehazing has been studied extensively in the field of computer vision recently \cite{14,23,28,30,36,38,43,44,47,contrast2022,frequencydehazing2022}.

Early methods for image dehazing \cite{scatterdehaze1,scatterdehaze2} are primarily developed based on the atmospheric scattering model (ASM) \cite{ASM} to establish the correspondence between hazy images and hazy-free images. This model can be described by \cref{eq:ASM} below:
\begin{equation}
\setlength\abovedisplayskip{2pt}%shrink space
\setlength\belowdisplayskip{2pt}
  I(x) = J(x)t(x) + A(1-t(x)).
  \label{eq:ASM}
\end{equation}
Here $I$ and $J$ represent the hazy image and its clear counterpart, respectively; $x$ indicates the pixel position; $A$ denotes the global atmosphere light; $t(x)$ is the transmission map, which is determined by the atmosphere scattering parameter $\beta$ and the scene depth $d(x)$ as follows:
\begin{equation}
\setlength\abovedisplayskip{3pt}%shrink space
\setlength\belowdisplayskip{3pt}
  t(x)=e^{-\beta d(x)}.
  \label{eq:transmission map}
\end{equation}

It is clear that assuming the validity of ASM, image dehazing boils down to estimating $t(x)$ and $A$ \cite{scatterdehaze2,23,31}. Unfortunately, this model is only applicable to idealized homogeneous haze. As such, ASM based methods cannot handle non-homogeneous dehazing tasks.

%In order to obtain high-fidelity hazy-free images, many methods\cite{scatterdehaze2,23,31} attempt to precisely estimate $t(x)$ and $A$, then invert Eq. \eqref{eq:ASM} to 

%by assuming that the scene depth is highly correlated to the density of haze. However, this assumption is applicable only for hazy images with simple or homogeneous haze distribution, thus ASM based methods cannot be applied to non-homogeneous dehazing task. 

In recent several years, inspired by its remarkable success for classification, object detection and other vision tasks \cite{GAN,Res2Net,objectreview}, deep learning has also been brought to bear upon single image dehazing \cite{fusion,ntire2021,ntire2020,ntire2019,ntire2018,dehazing-deraining,contrast2022,frequencydehazing2022,Liu_2019_ICCV,Liu_2022_ITS}. In principle, with end-to-end supervised training, deep learning based dehazing methods are no longer confined by the ASM framework. Indeed, they are shown to be able to handle complex and non-homogeneous hazy images to a certain extent \cite{ntire2021,ntire2020}.

%Overall, among various deep learning based dehazing methods, CNNs are used to replace ASM to estimate the feature mapping between hazy and clean images. Considering the strong capability of CNN, these methods are no longer limited to homogeneous dehazing challenges but can dehaze complex and non-homogeneous hazy images to a certain extent\cite{ntire2021,ntire2020}. 

Deep learning based dehazing methods often rely on the availability of large training data. However, it is very difficult and even impossible to acquire big volumes of image pairs in the real world \cite{DW-GAN}. The lack of sufficient training data has become a hindrance to the development of non-homogeneous dehazing methods. To cope with limited training data and alleviate over-fitting, some recent methods \cite{ntire2020,ntire2021,DW-GAN} resort to pretrained models, e.g., Res2Net pertrained on ImageNet \cite{Res2Net,Imagenet}, for transfer learning \cite{transfer-learning1}. However, these methods do not take advantage of the state-of-the art models, such as Vision Transformer \cite{VisionTransformer} and its follow-ups \cite{ViT-followup1,ViT-followup2}, Swin Transformer \cite{Swin-T} and ConvNeXt \cite{convnext,convnext2}, which achieve superior performance compared to Res2Net on ImageNet. As such, there is a potential to improve the dehazing performance by utilizing more powerful pretrained models.

%Considering the importance of the pretrained model to the whole network, utilizing more powerful pretrained models is essential to improve the dehazing performance.

Besides, most existing methods are incapable of recovering high-frequency components, such as edges and fine textures. DW-GAN \cite{DW-GAN} leverages discrete wavelet transform (DWT) to extract high-frequency features in the downsampling phase and passes them through the upsampling process. However, this method still has difficulty in dealing with severe hazy areas.

Therefore, two problems need to be addressed in order to realize high-quality dehazing. 1) Including those released by NTIRE \cite{nh-haze2020,densehaze2019}, most datasets for non-homogeneous dehazing are small-scale ones, which are not sufficient for CNN-based models to learn the mapping between hazy images and its corresponding hazy-free images. Although adopting a pretrained model can alleviate the over-fitting problem caused by limited training data, the network should be designed carefully to maximize its ability to acquire prior knowledge. 2) The complicated and non-uniform haze patterns pose significant challenges to image restoration, especially regarding dense haze areas \cite{DW-GAN,ntire2021}.

%The haze patterns in most datasets are complicated and non-uniform. Thus current methods usually fail to recover the hazy image, especially the severe haze area \cite{DW-GAN,ntire2021}.

In consideration of the aforementioned two problems, we propose a two-branch generative adversarial network, with each branch designed to address one problem.
 The first branch aims to learn the color and structure mapping from hazy to clean images. It consists of three DWT blocks and three FFC \cite{FFC,lama} residual blocks. DWT blocks are used to acquire high-frequency knowledge and structure details \cite{DW-GAN,WSAMF-Net,Liu_2020_ECCVW}, and FFC residual blocks provide a wide mixed receptive field that covers an entire image by simultaneously exploiting spectral and spatial information \cite{lama,glama}. Due to the larger receptive field, FFC residual blocks enable the encoder to capitalize on the global context and thus improve the perceptual quality of the recovered image, which is especially crucial for high-resolution non-homogeneous dehazing. The second branch serves the purpose of transfer-learning \cite{transfer-learning1,transfer-learning2,transfer-learning3}. Specifically, we employ the first three layers of a pretrained ConvNeXt to build the encoder due to its outstanding classification performance. In general, pretrained models can successfully adapt to different tasks than what they were originally trained for \cite{ntire2021}. Compared to Res2Net, ConvNeXt performs much better on ImageNet with the incorporation of several key components of vision transformer and the preservation of the advantage of convolutional network \cite{convnext,VisionTransformer}. For the decoder of the second branch, pixel-shuffle layers and channel/pixel wise attention blocks \cite{pixel-shuffle-attention} are employed to gradually recover the image to its initial resolution. Then, a fusion operation \cite{ntire2021} is utilized to aggregate the outputs of the two branches. Finally, a discriminator guided by the adversarial loss \cite{DW-GAN} is introduced to ensure the perceptual quality of the final reconstruction.

The main contributions of our work are as follows: 1) We introduce FFC residual blocks and combine them with 2D discrete wavelet transform to tackle non-homogeneous dehazing. 2) We employ the pretrained ConvNeXt model to leverage the prior knowledge to cope with the small-scale dataset problem and verify the effectiveness of this approach. 3) We conduct extensive experiments and  ablation studies to justify the overall design and demonstrate its competitive performance.

%-------------------------------------------------------------------------
\section{Related Works}
\label{sec:rela}
\textbf{Single Image Dehazing.} Single image dehazing is a challenging task in computer vision and image processing, as it involves removing the unwanted atmospheric haze from a single input image. Over the years, several effective methods that can be divided into two main categories, $i.e.,$ physical-based methods and CNN-based solutions, have been proposed to tackle this problem.  Physical-based methods mainly depend on ASM \cite{ASM,ASM2} and the hand-crafted priors, like dark channel prior \cite{23}, color attenuation prior \cite{attenuation-prior}, non-local prior  \cite{nolocal-piror}. However, owing to the limited applicability of the underlying assumptions, physical-based methods tend to be not very robust.

With the rapid advancement of deep learning, the past few years witnessed its wide applications in single image dehazing. Early deep learning based methods still utilize ASM. For instance, DehazeNet \cite{14} designs CNN model to estimate the medium transmission map, then uses it to obtain a dehazed image via ASM. Later, AOD-Net \cite{28} estimates the atmospheric light and transmission map simultaneously to generate the recovered image. Recently, various deep learning models have been proposed to directly map hazy images to their clean counterparts without resorting to ASM. For example, GCANet \cite{dehazing-deraining} introduces a gated context aggregation network to remove the grid artifacts and fuse the feature representations of different levels. Qin et al.\cite{36} propose FFA-net which handles different features and pixel regions adaptively in order to enhance flexibility via groups of channel attention and pixel attention mechanisms. MSBDN \cite{fusion} employs a multi-scale boosted decoder to gradually recover the dehazed images. Most previous works assume a homogeneous distribution of haze, which is often not representative of real-world scenarios and can lead to significant performance degradation in scenes with dense or non-homogeneous haze. To handle real-world hazy images, Trident Dehazing Network (TDN) \cite{ntire2020}, which consists of details refinement sub-net, encoder-decoder sub-net and haze density map generation sub-net, has been proposed in the NTIRE 2020 NonHomogeneous dehazing challenge. In the NTIRE 2021 NonHomogeneous dehazing challenge \cite{ntire2021}, TDN is surpassed by DW-GAN \cite{DW-GAN} which employs DWT to extract low-frequency and high-frequency features, which are beneficial to the dehazing task. Besides, \cite{contrast2022} introduces a novel dehazing approach based on contrastive learning that leverages both positive and negative image information. All the aforementioned methods, except for DW-GAN, are spatial-domain-centric dehazing methods, which do not directly exploit the characteristics of haze degradation in the frequency domain \cite{frequencydehazing2022}.% resulting in sub-optimal performance on non-homogenous dehazing task.

\textbf{Frequency Domain Learning.} In recent years, there has been a growing research attention to frequency domain learning in pursuit of effective spectral features. Specifically, \cite{wavelete-domain} exploits wavelet-based representations that facilitate high-resolution image restoration. \cite{wavelete-gan} decomposes images into low-frequency and high-frequency bands via discrete wavelet transform and extract features for each band. In addition, Chi et al. \cite{FFC} propose FFC as a non-local operation unit that concurrently enlarges the receptive field by learning spatial and spectral features. Subsequently, \cite{lama} achieves excellent performance in the large mask inpainting task by employing FFC as the main convolution operation and the proposed method exhibits strong generalization abilities. \cite{frequencydehazing2022} unveils the connection between haze degradation and the frequency property and designs a dual-branch network that guide the dehazing process spatially and spectrally. 

The above-mentioned methods have demonstrated that frequency domain information can be leveraged to significantly enhance the performance of image restoration methods. As such, We borrow the idea of wavelet-based decomposition and incorporate non-local FFC operation units into our deep learning network architecture. By this, our model can effectively utilize spectral information while maintaining meaningful texture details and ensuring the consistent structures of the final recovered high-resolution image. We shall verify the superiority of our proposed model over the existing ones through quantitative evaluation.

\textbf{Transfer Learning.} Transfer learning aims to enhance the capability of target models on specific tasks by leveraging the knowledge acquired from related but distinct tasks, which weakens the need for large volumes of data on target domain \cite{transfer-learning1,transfer-learning2,transfer-learning3}. Some existing methods employ substantial prior knowledge obtained through pre-training on ImageNet to assist image restoration tasks. As an example, the champions of both NTIRE dehazing challenges in 2020 and 2021 \cite{ntire2020,ntire2021} utilize the pretrained Res2Net model as the fundamental block for knowledge transfer, successfully alleviating the over-fitting problem caused by small-scale training datasets. Despite its impressive capabilities, Res2Net exhibits limited efficacy in dehazing, particularly for high-resolution images with non-uniform haze. We intend to substantiate this through our experimental analysis. Furthermore, it is worth noting that even in the classification area, the performance of CNNs, like Res2Net, has been overshadowed by that of Vision Transformers \cite{VisionTransformer}.  Liu et al. \cite{convnext} redesign the standard ResNet \cite{resnet} by mimicing Vision Transformers through the incorporation of several key components that contribute to the remarkable performance of transformers.
 The resulting architecture, named ConvNeXt, outperforms Swin Transformers \cite{Swin-T} and challenges the widely held belief that Vision Transformers are more accurate and efficient than CNNs. However, to the best of our knowledge, ConvNeXt-based image restoration is still a largely unexplored territory. %To the best of our knowledge, our work firstly studies ConvNeXt-based pre-training for image dehazing. 

\section{Proposed Method}
\label{sec:Meth}
\begin{figure*}[t] 
\centering \includegraphics[width=0.93\linewidth]{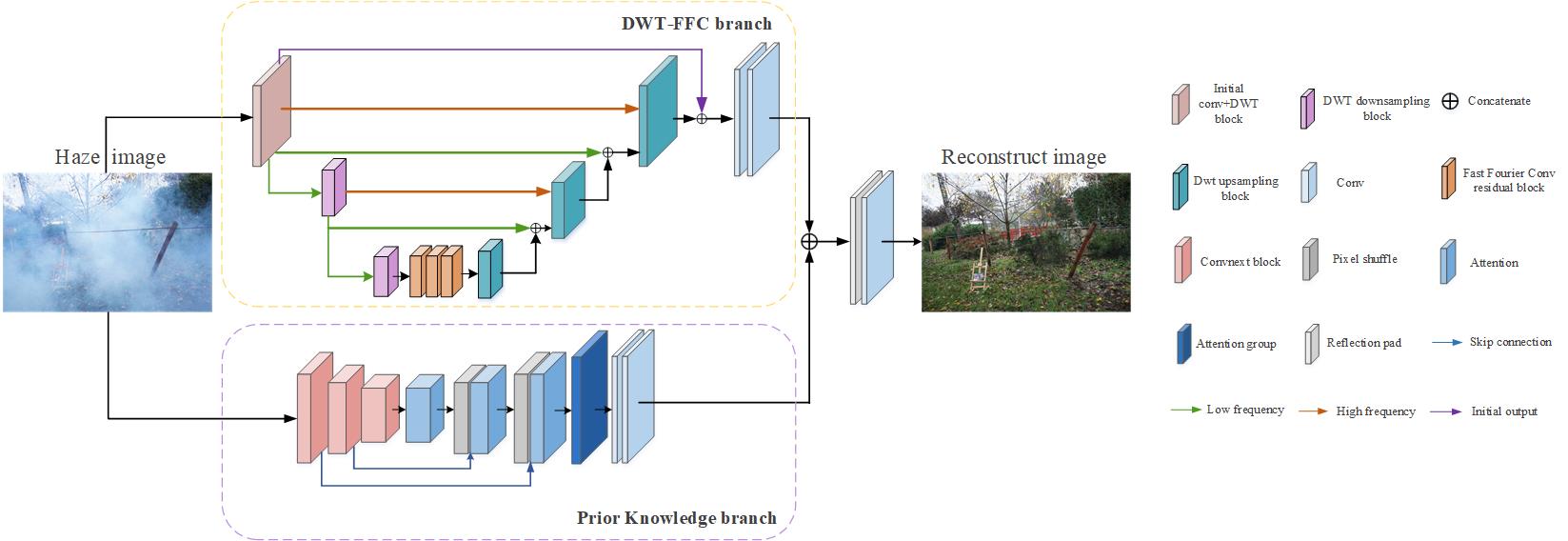} \caption{The network structure of our proposed method. } \label{fig::model} 
\end{figure*}

Within this part, we first describe the details of our proposed network (shown as \cref{fig::model}) based on DWT, FFC residual block and pretrained ConvNeXt parameters. Then, we introduce DWT and FFC residual blocks in detail, and analyze their significance to the whole network, respectively. Finally, we discuss the loss functions utilized to train our model.
\subsection{Network Framework}
\label{sec:network}
Many methods with two-branches have shown great success in NTIRE 2020 and 2021 NonHomogeneous dehazing challenge \cite{ntire2020,ntire2021}. In observing that, we design a two branch neural network (shown as \cref{fig::model}).

\begin{figure}[t] 
\centering \includegraphics[width=0.68\linewidth]{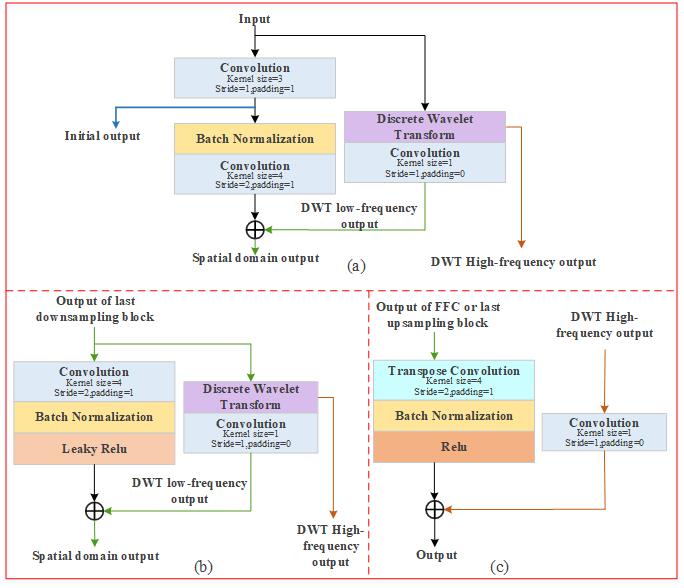} 
\caption{The illustration of Connections between Conventional Convolution and DWT in DWT-FFC branch. Green lines represent DWT low-frequency features, brown lines denote DWT high-frequency features.} \label{fig::DWT} 
\end{figure}

\begin{figure}[t] 
\centering \includegraphics[width=0.8\linewidth]{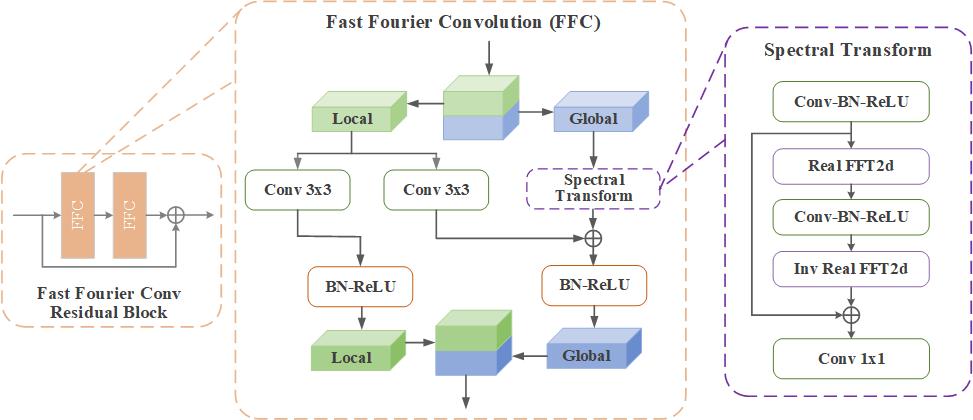} \caption{The network structure of FFC residual block. A FFC residual block has two FFC units, each unit contains one spectral transform.} \label{fig::FFC} 
\end{figure}

\textbf{DWT-FFC branch.} Inspired by \cite{ntire2021}, we construct our DWT-FFC frequency branch as a encoder-decoder network to learn the feature mapping between hazy images and clear images, and we leverage massive skip connections at each feature scale. Besides conventional convolution techniques, we adopt discrete wavelet transform (DWT) to achieve feature extraction. By using DWT, both high-frequency and low-frequency features can be detected by our model (\cref{sec:DWT} provides detailed explanation). As indicated in \cref{fig::DWT}, low-frequency representations are concatenated with common convolution output, whereas high-frequency features are transferred to the up-sampling module to recover the hazy-free image gradually. In order to make the reconstructed image more realistic and perceptual, as shown in \cref{fig::FFC}, we propose to utilize fast Fourier convolution residual blocks (FFC, details can be found in \cref{sec:FFC}) to utilize the spatial and spectral information for dehazing. 

%

%\begin{figure*}[t] 
%\centering \includegraphics[width=0.75\linewidth]{FFC block.pdf} %\caption{The network structure of FFC residual block.} \label{fig::FFC} 
%\end{figure*}

However, the model sorely with DWT-FFC frequency branch cannot achieve plausible performance for non-homogeneous challenge due to the lack of large volumes of training data. Therefore, we further introduce the second branch with a strong network pretrained on large datasets to acquire prior knowledge.

\textbf{Prior Knowledge Branch.} Our second branch, prior knowledge branch, is designed to bring additional information from large dataset image classification task to the current non-homogeneous dehazing challenge. Specifically, we design this branch similar to the U-Net \cite{U-Net}, and utilize the ImagNet pretrained ConvNeXt as the backbone of the encoder. We only adopt the first three stages of ConvNeXt. In this way, the model with the pretrained weights can perform better than that with random initial parameters. For the decoder, inspired by \cite{36}, several upsampling layers are employed and each upsampling layer contains a pixel-shuffle block and a attention module. Here, pixel-shuffle blocks are introduced to decrease the computational burden and make the size of the feature maps gradually recover to the original resolution, attention blocks enable our model to identify the dynamic hazy patterns. Moreover, as shown in \cref{fig::FFC}, skip connection technique is used to enhance the generalization ability. Finally, final recovered clear images are generated by combining the outputs of each branch via a simple and effective fusion \cite{two-branch}.
\subsection{Discrete Wavelet Transform}
\label{sec:DWT}
2D discrete wavelet transform has one low-pass filter $(f_{LL} )$ and three high-pass filters $(f_{LH}, f_{HL}, f_{HH} )$, and these filters are equivalent to convolution operations with fixed parameters and stride 2. For example, in Haar DWT introduced in our method, $f_{LL}=\left (\begin{matrix}1 & 1 \\ 1 & 1 \\ \end{matrix}\right)$, $f_{LH}=\left (\begin{matrix}-1 & -1 \\ 1 & 1 \\ \end{matrix}\right)$, $f_{HL}=\left (\begin{matrix}-1 & 1 \\ -1 & 1 \\ \end{matrix}\right)$, $f_{HH}=\left (\begin{matrix}1 & -1 \\ -1 & 1 \\ \end{matrix}\right)$. After convolving with each filter, the input can be decomposed into four sub-components, $i.e., x_{LL}, x_{LH}, x_{HL}$ and $x_{HH}$. Specifically, $x_{LL}$ can be expressed as \cref{eq:x_LL}, where $i$ and $j$ donates the index of pixel, and the expressions of the other three sub-bands are similar to that of $x_{LL}$.
\begin{equation}
\setlength\abovedisplayskip{2pt}%shrink space
\setlength\belowdisplayskip{2pt}
\begin{split}
  x_{LL} &= x(2i-1,2j-1)+x(2i-1,2j)\\
  &+x(2i,2j-1)+x(2i,2j)
  \label{eq:x_LL}
\end{split}
\end{equation}
Besides, we combine $x_{LL}$ with conventional convolution outputs at each feature scale so that our proposed model can learn spatial and frequency information.
\subsection{Fast Fourier Convolution}
\label{sec:FFC}
Fast Fourier convolution, based on the channel-wise fast Fourier transform, has a large receptive field as the entire image or feature map. FFC splits the input into two parallel branches: $i)\ local\ branch$ applies two conventional convolutions in parallel, and $ii)\ global\ branch$ simultaneously utilizes one conventional convolution and one spectral transform to account for the global context. Specifically, there are three steps in spectral transform:

a) For an input tensor, the $Real\ FFT2d$ is applied and the real and imaginary parts are concatenated:

\textit{Real FFT2d}: $\mathbb{R}^{H\times W \times C} \rightarrow \mathbb{C}^{H\times \frac {W}{2} \times C}$,

\textit{ComplexToReal}: $\mathbb{C}^{H\times \frac {W}{2} \times C} \rightarrow \mathbb{R}^{H\times \frac {W}{2} \times {2C}}$;

b) ReLU activation, batch normalization and convolution operation are applied in the frequency domain:

\textit{ReLU $\cup$ BN $\cup$ Conv1$\times$1}:$\mathbb{R}^{H\times \frac {W}{2} \times {2C}}\rightarrow \mathbb{R}^{H\times \frac {W}{2} \times {2C}}$;

c) The inverse FFT transform is applied to recover the spatial structure:

\textit{RealToComplex}: $\mathbb{R}^{H\times \frac {W}{2} \times {2C}} \rightarrow \mathbb{C}^{H\times \frac {W}{2} \times {C}}$,

\textit{Inverse Real FFT2d}: $\mathbb{C}^{H\times \frac {W}{2} \times {C}} \rightarrow \mathbb{R}^{H\times W \times {C}}$.

As shown in \cref{fig::FFC}, FFC leverages the fusion operation to combine the outputs of the local and global branch. Finally, two FFC units form a FFC residual block by connecting the input to the output. In our proposed method, we adopt three FFC residual blocks in total to make our network remove the haze from the perspective of considering the entire image or feature map.
\subsection{Loss Function}
\label{sec:loss}
 We define the ground truth image as $I^{gt}$, and we denote the hazy image and the dehazed image as $I^{hazy}$ and $\Tilde{I}$, respectively. We utilize G and D to represent our proposed method and the discriminator.

\textbf{Smooth L1 Loss.} The smooth L1 Loss can be calculated using \cref{eq:L1-loss} and \cref{eq:f}, where $N$ denotes the total number of pixels, $I_{i}^{gt}(x)$ and $\Tilde{I}_{i}(x)$ represent the strength of the pixel $x$ in the i-th channel of the ground truth image and of the dehazed image.
\begin{equation}
\setlength\abovedisplayskip{0.1pt}%shrink space
\setlength\belowdisplayskip{0.1pt}
  \mathcal{L}_{\rm smooth-L1}= \frac{1}{N}\sum_{x=1}^N\sum_{i=1}^3f(I_{i}^{gt}(x)-\Tilde{I}_{i}(x))
  \label{eq:L1-loss}
\end{equation}
where 
\begin{equation}
\setlength\abovedisplayskip{0.1pt}%shrink space
\setlength\belowdisplayskip{0.1pt}
  f(\gamma)=
  \begin{cases}
      0.5\gamma^2 & {\rm if  \left| \gamma \right| < 1}\\
      \left| \gamma \right|-0.5& {\rm otherwise}
  \end{cases}
  \label{eq:f}
  \end{equation}
  
\textbf{Perceptual Loss.} We introduce the VGG-16 \cite{VGG-16} pre-trained on ImageNet as the loss network $\phi$. The perceptual loss can be defined as \cref{eq:PL}:
\begin{equation}
\setlength\abovedisplayskip{0pt}%shrink space
\setlength\belowdisplayskip{0pt}
  \mathcal{L}_{\rm PL}= \sum_{j=1}^3 \frac{1}{C_jH_jW_j}\left\| \phi_j(G(I_{i}^{hazy}))-\phi_j(I_{i}^{gt}) \right\|_2^2
  \label{eq:PL}
\end{equation}
where $\phi_j$ denotes the activation of the j-th layer in the backbone network, and $C_j$, $W_j$ and $H_j$ represent the channel, width and height of the corresponding feature map.
The ablation study can demonstrated that the perceptual loss appears to be crucial to our proposed method.

\textbf{MS-SSIM Loss.} Multi-scale Structure similarity (MS-SSIM) \cite{DW-GAN} is introduced in our loss function. We first calculate the SSIM for pixel $i$ using \cref{eq:SSIM}:
\begin{equation}
\setlength\abovedisplayskip{0pt}%shrink space
\setlength\belowdisplayskip{0pt}
\begin{split}
   SSIM(i)&= \frac{2\mu_D\mu_C+T_1}{\mu_D^2+\mu_C^2+T_1}\cdot\frac{2\sigma_{DC}+T_2}{\sigma_D^2+\sigma_C^2+T_2}\\
  &=l(i)\cdot s(i)
  \label{eq:SSIM}
\end{split}  
\end{equation}
where $T_1$ and $T_2$ denote two small constants, $D$ and $C$ are two fixed size windows centered at current pixel in the reconstructed image and in the clear image, respectively. After applying Gaussian filters, we can compute the means $\mu_D$, $\mu_C$, standard deviations $\sigma_D$, $\sigma_C$ and covariance $\sigma_{DC}$. The MS-SSIM loss is described as \cref{eq:ssim-loss}, where $S$ represents the total number of scales, $\alpha$ and $\beta$ are default parameters.
\begin{equation}
\setlength\abovedisplayskip{0pt}%shrink space
\setlength\belowdisplayskip{0pt}
\mathcal{L}_{\rm MS-SSIM}=1-\prod_{s=1}^S (l^{\alpha}(i) \cdot cs_s^{\beta_s}(i))
\label{eq:ssim-loss}  
\end{equation}

\textbf{Adversarial Loss.} The adversarial loss is calculated as \cref{eq:adv-loss}, where $D(G(I^{hazy}))$ represents the possibility that the recovered image is considered as a ground truth image by the discriminator \cite{GAN}.
\begin{equation}
\setlength\abovedisplayskip{0pt}%shrink space
\setlength\belowdisplayskip{0pt}
  \mathcal{L}_{\rm adv}=1-\sum_{s=1}^N -log D(G(I^{hazy}))
  \label{eq:adv-loss}  
\end{equation}

\textbf{Total Loss.} The total loss used to supervise the training of our proposed method is shown as \cref{eq:total-loss} :
\begin{equation}
\setlength\abovedisplayskip{2pt}%shrink space
\setlength\belowdisplayskip{2pt}
  \mathcal{L}_{\rm total}= \mathcal{L}_{\rm smooth-L1}+ \alpha\mathcal{L}_{\rm MS-SSIM}+\beta\mathcal{L}_{\rm PL}+ \gamma\mathcal{L}_{\rm adv} 
  \label{eq:total-loss}
\end{equation}
where $\alpha=0.2$, $\beta=0.01$, $\gamma=0.0005$ are the hyper-parameters for each loss function.
\begin{table*}[t]
\setlength\abovedisplayskip{0pt}%shrink space
\setlength\belowdisplayskip{0pt}
  \centering
  \renewcommand{\arraystretch}{0.9}
  \begin{tabular}{ccccccc}
    \toprule
    Methods & $L_1$ & $L_{PL}$ & $L_{ssim}$ & $L_{D}$ & PSNR & SSIM  \\
    \midrule
    (1) Single DWT branch & \checkmark & \checkmark & \checkmark & \checkmark & 18.35 & 0.6491 \\ 
    (2) Single FFC branch & \checkmark & \checkmark & \checkmark & \checkmark & 19.03 & 0.6834 \\ 
    (3) Single DWT-FFC branch & \checkmark & \checkmark & \checkmark & \checkmark & 19.38 & 0.6991 \\ 
    (4) Prior knowledge branch & \checkmark & \checkmark & \checkmark & \checkmark & 20.35 & 0.7161 \\ 
    (5) DWT-FFC branch+Res2Net  & \checkmark & \checkmark & \checkmark & \checkmark & 21.67 & 0.7301 \\ 
    (6) Ours & \checkmark & \checkmark & \checkmark & \checkmark & \textbf{22.20} & \textbf{0.7458} \\
    \midrule 
    (7) ours & \checkmark & \checkmark & \checkmark & & 22.07 & 0.7427 \\ 
    (8) ours & \checkmark & \checkmark & & & 21.81 & 0.7344 \\
    (9) ours & \checkmark & & & & 21.71 & 0.7299 \\ 
    \bottomrule
  \end{tabular}
  \captionsetup{justification=centering}
  \caption{Results of Ablation Study. The first raw is to study the necessity of each component of the network; The second raw is to illustrate the rationality of loss function used for training. Figures in \textbf{bold} denote the best results.}
  \label{tab:ablation-study}
\end{table*}

\begin{figure*}[h]
\centering
\begin{minipage}[h]{0.49\linewidth}
\centering
\includegraphics[width=0.976\linewidth]{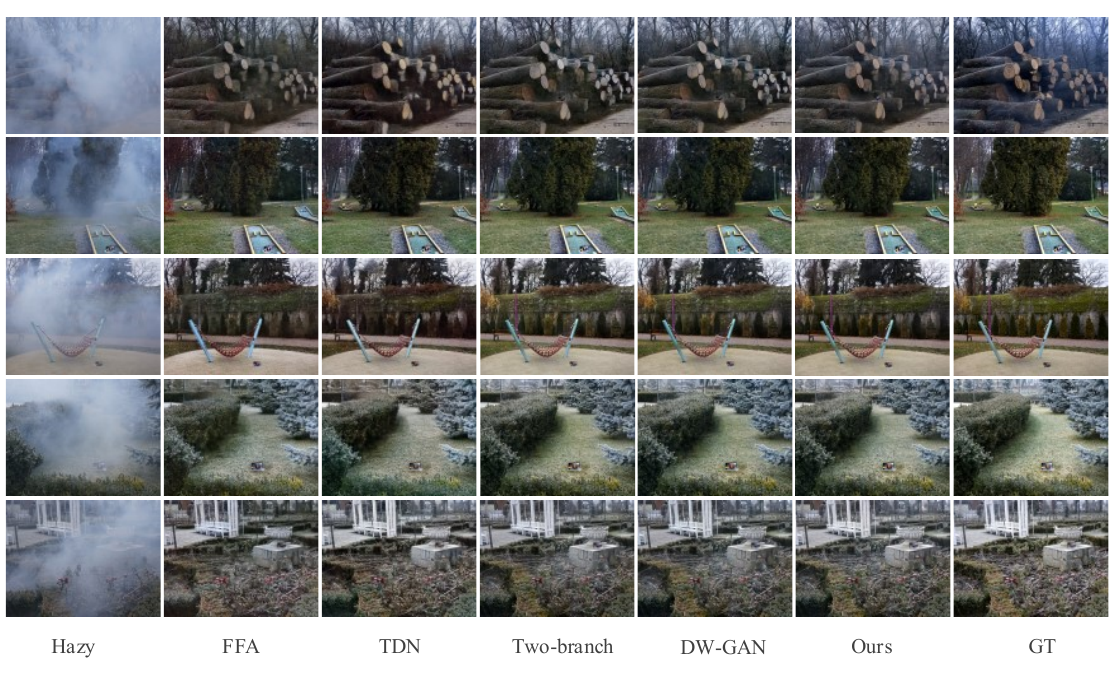} \caption{Comparison results on NH-HAZE.} \label{fig::comparison-ntire20}
\end{minipage}
\begin{minipage}[h]{0.49\linewidth}
\centering
\includegraphics[width=\linewidth]{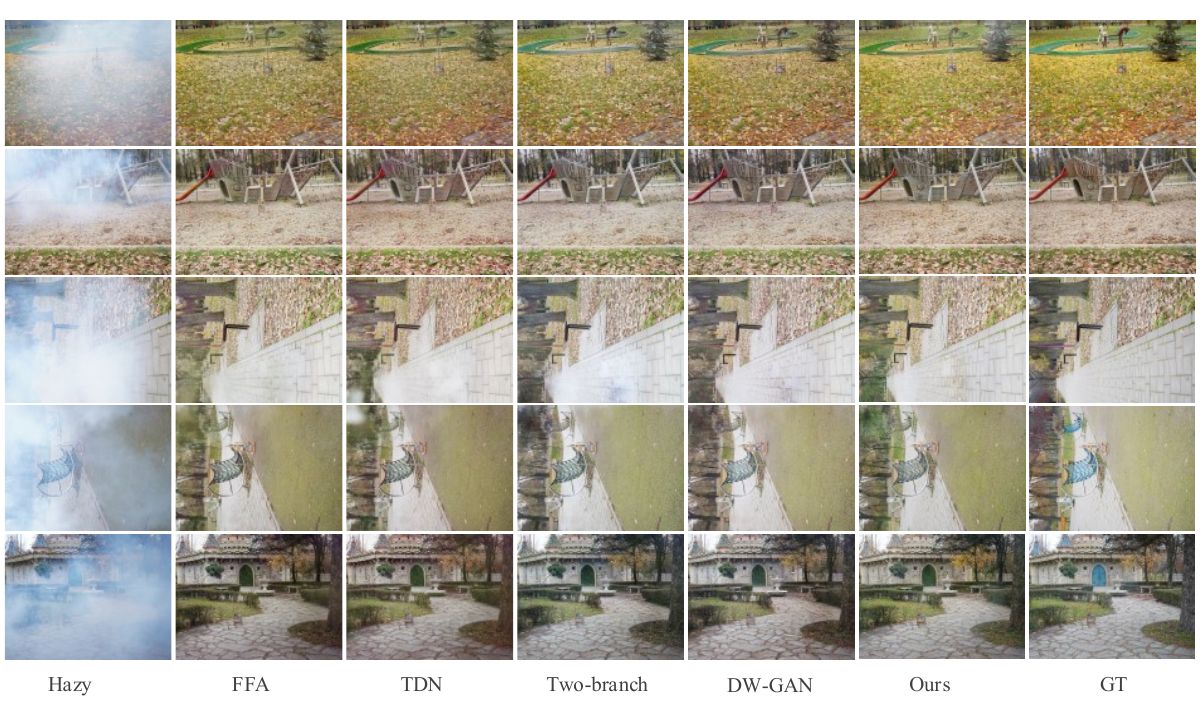} \caption{Comparison results on NH-HAZE2.} \label{fig::comparison-nhaze2} 
\end{minipage}

\end{figure*}

\begin{figure}[t]
\centering
\begin{minipage}[h]{0.50\linewidth}
\centering
\includegraphics[width=\linewidth]{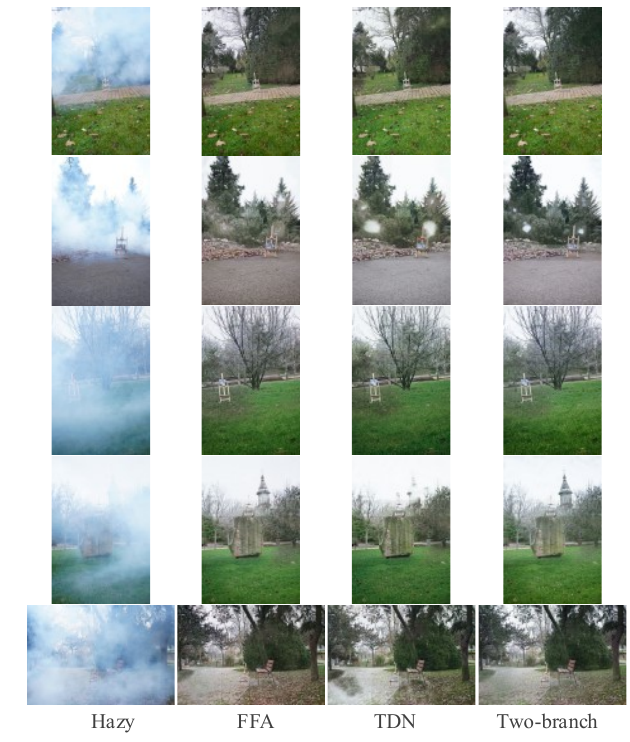} 
\end{minipage}
\begin{minipage}[h]{0.39\linewidth}
\centering
\includegraphics[width=\linewidth]{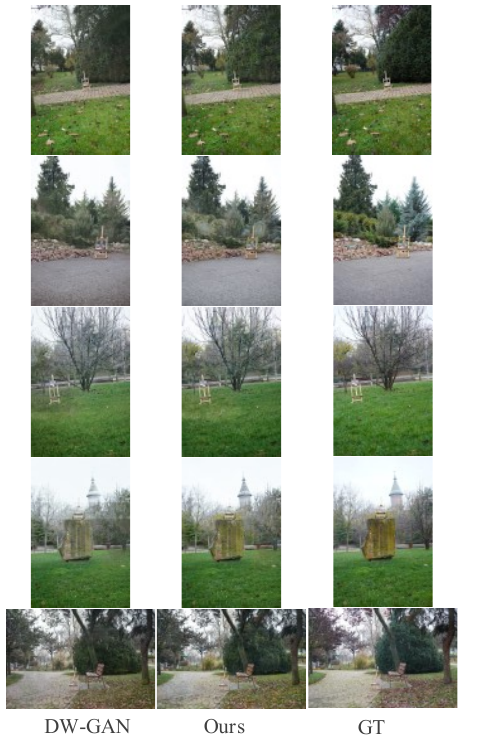}  
\end{minipage}
\caption{Comparison results on HD-NH-HAZE. Our results are nearly closed to the ground truth and are highly authentic with little noticeable dissonance.}
\label{fig::comparison-ntire23}
\end{figure}

\begin{figure}[t] 
\centering \includegraphics[width=\linewidth]{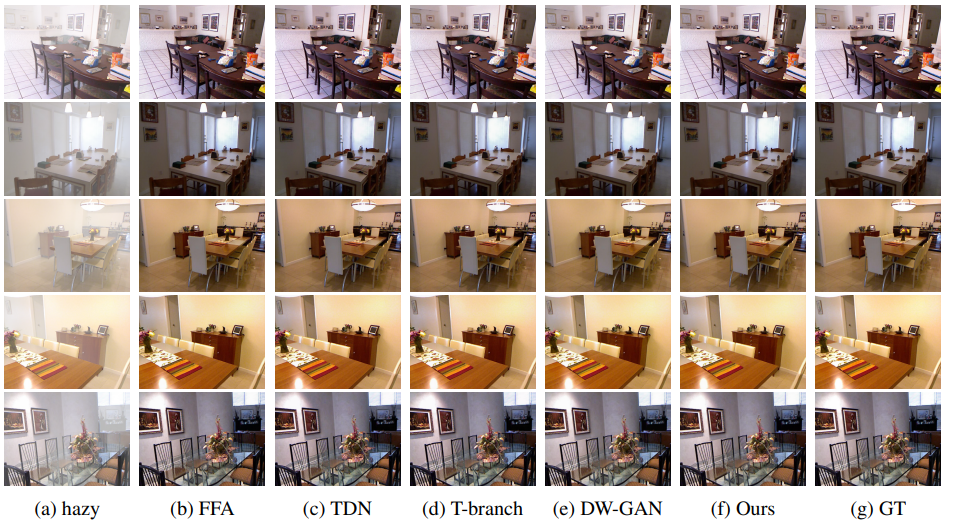} \caption{Comparison results on ITS. All methods generate visually pleasing dehazed images.} \label{fig::comparison-its} 
\end{figure}

\begin{table*}[t]
  \centering
  \resizebox{\linewidth}{!}{
  \renewcommand{\arraystretch}{1}
  \begin{tabular}{ccccccccccccc}
    \toprule
    \multirow{2}{*}{Methods} &\multicolumn{2}{c}{ITS} &\multicolumn{2}{c}{NTIRE20} &\multicolumn{2}{c}{NTIRE21} &\multicolumn{2}{c}{HD-NH-HAZE} 
    &\multicolumn{2}{c}{Combined Dataset} \\
    \cline{2-11}  
    %\cline{4-5} \cline{6-7} \cline{8-9} \cline{10-11}\cline{12-13}
    & PSNR & SSIM & PSNR & SSIM & PSNR & SSIM & PSNR & SSIM & PSNR & SSIM \\
    \midrule
    FFA &36.55 & 0.9888 & 19.50 & 0.6441 & 20.56 & 0.8106 & 20.23 & 0.7103 & 20.43 & 0.7112 \\ 
    TDN &34.87 & 0.9804 &20.73 & 0.6727 &20.44 & 0.8014 & 20.06 & 0.7132 & 20.30 & 0.7140\\ 
    Two-branch &\uuline{37.76}\par & \uuline{0.9905}\par &21.55 & 0.7149 & 21.75 & 0.8277 & 21.15 & 0.7323 & 21.39 & 0.7327\\ 
    DW-GAN &36.33 & 0.9866 &\uuline{21.69}\par & \uuline{0.7161}\par & \uuline{22.13}\par & \uuline{0.8314} & \uuline{21.52}\par & \uuline{0.7325}\par & \uuline{21.68}\par & \uuline{0.7371}\par\\ 
    ours &\textbf{37.86} & \textbf{0.9907} &\textbf{22.64} & \textbf{0.7298} &\textbf{22.82} & \textbf{0.8738} & \textbf{22.20} & \textbf{0.7458} & \textbf{22.26} & \textbf{0.7469} \\ 
    \bottomrule
  \end{tabular}}
  \captionsetup{justification=centering}
  \caption{Quantitative comparisons over ITS, NH-HAZE, NH-HAZE2, HD-NH-HAZE and Combined Dataset. Figures in \textbf{bold} denote the best results, and results with \uuline{double underline} represent the second best.}
  \label{tab:quantative-comparison-results}
\end{table*}

\section{Experiments}
\label{sec:experiments}
In this section, we first introduce the datasets. Then, we discuss the experiment settings and evaluation criteria. Besides, we present the ablation study for our network. Moreover, we compare our method with state-of-art dehazing methods quantitatively and qualitatively. Finally, we introduce our dehazing results in NTIRE $2023$ High-Resolution NonHomogeneous Dehazing Challenge.
\subsection{Datasets}
\label{sec:datasets}
 We conduct experiments utilizing RESIDE (Indoor Training Set, ITS) \cite{RESIDE2019}, NH-HAZE \cite{nh-haze2020}, NH-HAZE$2$ \cite{ntire2021}, and HD-NH-HAZE \cite{ntire23}. ITS encompasses more than ten thousands of training samples, and its Synthetic Objective Testing Set (SOTS) can be used for testing. NH-HAZE contains a total of $55$ images. we employ the official testing data for model evaluation while the remaining are utilized for model training. NH-HAZE$2$ consists of a mere $25$ training data, $5$ validation data, and $5$ test data. Given the fact that both the validation and test data remain undisclosed, we extract the first $20$ images from the training data for training, and utilize the remaining $5$ images as testing samples. The HD-NH-HAZE dataset is characterized by dense and non-uniformly distributed hazy scenes. It comprises $40$ training data, $5$ validation data, and $5$ testing data, each with a resolution of $6000\times4000$. Considering the validation and testing sets are not publicly available at present, we utilize images $1$-$35$ as training data while images $36$-$40$ are reserved for evaluation purposes. Besides, we have utilized gamma correction to build a combined dataset with $120$ image pairs based on NH-HAZE, NH-HAZE$2$ and HD-NH-HAZE dataset (More details can be found in \cref{sec:ntire23-chanllenge}). We use the same $5$ images with HD-NH-HAZE for testing, and train models using the rest $115$ images.
\subsection{Experimental Settings}
\label{sec:experimental-settings}
To diversify the training data for our model, we incorporate random cropping of patches with a size of $384\times384$, accompanied by random rotation at $90$, $180$, or $270$ degrees, vertical flip and horizontal flip. The optimization process is facilitated by the Adam optimizer, with default values of $\beta_1$ and $\beta_2$ ($0.9$ and $0.999$, respectively). Our training method features a specialized decay strategy, which initiates with a learning rate of $e^{-4}$ and undergoes a decay of $0.5$ times at $3000$, $5000$, and $8000$ epochs, ultimately concluding after $10000$ epochs. The discriminator uses the same optimizer and training strategies. To execute the experiments, we utilize one RTX $2080$ Ti GPU. To undertake a quantitative assessment of our model efficacy, we employ two commonly used metrics, namely the Peak Signal to Noise Ratio (PSNR) and the Structural Similarity Index (SSIM).
Due to our limited GPU resource, our trained model cannot process an entire image from HD-NH-HAZE, we apply a block-based testing strategy to tackle this problem: 1) we first split the image into 9 blocks and the size of each block is $1600\times2432$; 2) we then output the dehazed results for each block using our trained weights; 3) we finally integrate these outputs into the final result of the same size of the hazy input by averaging the overlapping regions.
\subsection{Ablation Study}
\label{sec:ablation-study}
We conduct comprehensive ablation studies, which aim to establish the indispensability of each component in our proposed methodology. Consistent with the principles of ablation, we devise and construct six distinct networks, each featuring a unique combination of modules. These networks serve to demonstrate the relative significance of each module and underscore the importance of the individual components in our overall framework. (1) Single DWT branch: only using three DWT downsampling and three DWT upsampling without FFC. (2) Single FFC branch: solely utilizing FFC modules, without the inclusion of DWT downsampling blocks or high-frequency skip connections. (3) Single DWT-FFC branch: combining DWT downsampling and upsampling with FFC modules as a single branch. (4) Prior knowledge branch:  only using the transfer learning branch to restore defogging images. (5) DWT-FFC frequency branch and Res2Net: The only distinction between this particular network and our final methodology lies in the encoder of the transfer learning branch. (6) Ours: Two-branches consists of DWT-FFC frequency branch and ConvNext based prior knowledge branch. 

By comparing (3) to (1) and (2) in \cref{tab:ablation-study} respectively, we can conclude that both DWT and FFC can help improve the dehazing performace in terms of PSNR and SSIM, and even FFC plays a more prominent role than DWT. This observation is consistent with our discussion for DWT (\cref{sec:DWT}) and FFC (\cref{sec:FFC}). Moreover, we can observe that the performance of our proposed two-branch network is far superior to that of any single branch (by comparing (6) to (3) and (4), respectively). This finding validate our initial intention for designing dual-branch network. Finally, we compare our model to another two-branch network in which the pretrained Res2Net weights are introduced. The increased PSNR and SSIM between (6) and (5) in \cref{tab:ablation-study} can demonstrate the effectiveness of pretrained ConvNeXt weights. 
Besides, to verify the rationality of the total loss function used in training, we train our model with several combinations of loss functions introduced in \cref{sec:loss}. As shown (6)-(9) in \cref{tab:ablation-study}, our model can achieve the best performance using the loss function defined in \cref{eq:total-loss}, discarding any loss could worsen our model in terms of PSNR and SSIM.

\subsection{Comparisons with State-of-art Models}
In this section, we compare our proposed method with state-of-art models quantitatively and qualitatively. These methods include the winner solution in NTIRE 2020 NonHomogeneous Dehazing Challenge (TDN \cite{30}), the winner solution (DW-GAN) and a two branch dehazing method via ensemble learning (abbreviated as Two-branch in this paper) in NTIRE 2021 NonHomogeneous Dehazing Challenge and FFA \cite{36}. 

\begin{figure}[h] 
\setlength\abovedisplayskip{0pt}%shrink space
\setlength\belowdisplayskip{0pt}
\centering \includegraphics[width=\linewidth]{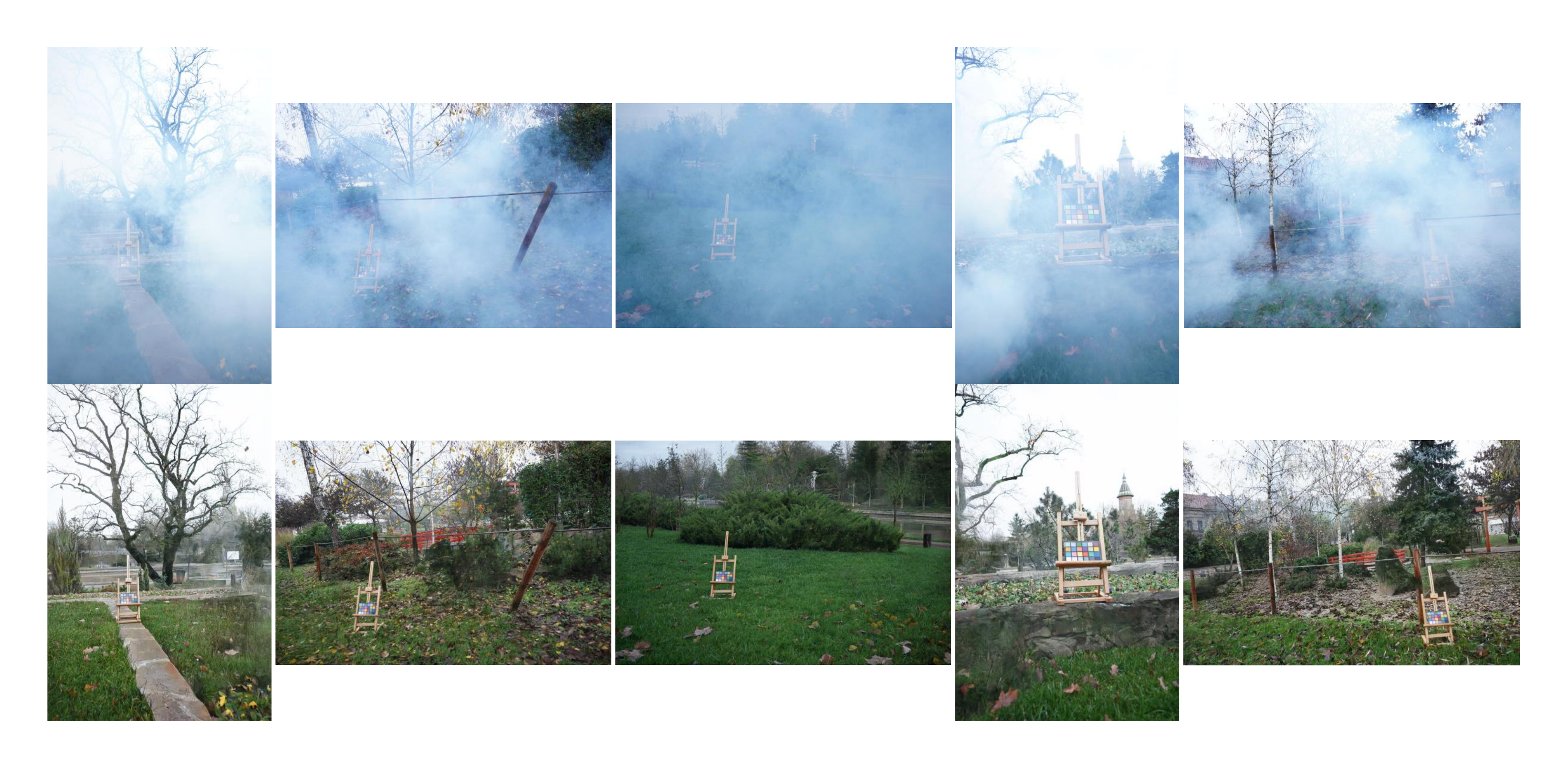} \caption{The test result of our method on NTIRE 2023 HR NonHomogeneous Dehazing Challenge.} \label{fig::test-ntire23} 
\end{figure}

\textbf{Quantitative Comparison Results.}
The comparison results are provided in \cref{tab:quantative-comparison-results}. The performance of our method greatly exceeds that of other models on all datasets. Specially, our model surpasses the second-ranked method by an average of $0.725$ dB for PSNR across these four non-homogeneous datasets. On the other hand, on ITS, the superority of our method compared to other models is not obvious (just $0.09$ dB higher than Two-branch in terms of PSNR). The reason for this phenomenon lies on the number of training data and complexity of hazy patterns. ITS is a large-scale synthetic dataset. Considering the availability of training data that exceeds 100 times than other four datasets, the contribution of the pretained ConvNeXt weights to the whole network has been attenuated. Besides, the haze in ITS is not as complicated as that in other datasets, the advantages of DWT and FFC residual blocks which are designed to address non-homogeneous and dense haze cannot be fully exploited. 

\textbf{Qualitative Comparison Results.}
In this section, we compare the performance of our model to other methods from the perceptual and visual standpoint. The comparison results are shown as \cref{fig::comparison-its}, \cref{fig::comparison-ntire20}, \cref{fig::comparison-nhaze2} and \cref{fig::comparison-ntire23}. For ITS, all methods can generate dehazed images that are visually pleasing and quite close to the ground truths. Large volumes of training data and relatively simple haze patterns enable these advanced models learn the feature and color mapping from hazy to clear images. For non-homogeneous datasets, the employment of distinct methods results in discernible visual discrepancies. Specifically, on NH-HAZE, blurred results generated by FFA and TDN indicate the limited ability of these two methods for non-homogeneous dehazing task, while Two-branch and DW-GAN methods suffer from a certain degree of chromatic and structural displacement. On NH-HAZE$2$, TDN performs comparably bad due to the presence of haze in the results, FFA, Two branch and DW-GAN methods can produce results that are highly close to the ground truth, but the unpleasing color shift still exists in some area, like the grass in the first test image and the road in the last image. With respect to HD-NH-HAZE dataset, there are several noticeable defects in the images processed by FFA, TDN and Two-branch method: the existence of white holes that are both meaningless and highly discordant. DW-GAN shows its image recovering ability based on extremely hazy images, however, the results contain several blurred local regions and exhibits noticeable fragmentation caused by block-based testing strategies. Generally, over all datasets, even though our method may produce artifacts or cannot remove haze on some small areas, like the trees in the first test image of NH-HAZE$2$ and in the last picture of HD-NH-HAZE, the color and structural details in vast majority of image regions are very nearly closed to the ground truth, and the results are highly authentic with little noticeable dissonance.

\subsection{NTIRE2023 HR NonHomogeneous Challenge}
\label{sec:ntire23-chanllenge}
\textbf{Combined Dataset based on NH-HAZE, NH-HAZE$2$ and HD-NH-HAZE.} The challenge only provides $40$ training samples, which are not abundant for the training process. In order to increase the number of training data and improve the generalization ability, we apply channel-wise gamma correction for NH-HAZE and NH-HAZE$2$, respectively. Specifically, in each channel of these two datasets, the gamma correction with a unique gamma parameter has been utilized to make the average intensity of this channel much closer to that of HD-NH-HAZE. By achieving this, we assume these two datasets share similiar haze patterns with HD-NH-HAZE and the network trained on the combined datasets may perform better than that trained sorely on HD-NH-HAZE. The results in \cref{tab:quantative-comparison-results} can indicate this tendency and verify the effectiveness of this pre-processing method.

\textbf{Performance on NTIRE 2023 High-Resolution NonHomogeneous Dehazing Challenge.}
According to the report \cite{ntire23}, our model is the winner of the challenge and is one of the top perceptual quality approaches in terms of PSNR (22.87) and SSIM (0.71). The test results of our model are illustrated in \cref{fig::test-ntire23}, which indicates our model features advanced capabilities in effectively removing haze, yielding visually appealing outputs with consistent structure.

\section{Conclusions}
In this paper, we propose a novel two branch network for high-resolution image dehazing. DWT-FFC frequency branch takes advantage of DWT to extract intricate high-frequency features while simultaneously employing FFC as a wide receptive field to enhance the perceptual quality of the output. The prior knowledge branch serves as providing supplementary information to address over-fitting on small-scale datasets, thereby improving the model generalization capability. Extensive empirical evaluations conclusively indicate that our model exhibits impressive performance on both synthetic datasets and real-world scenes. Moreover, our model surpasses the latest state-of-the-art techniques with superior fidelity and perceptual quality.

%%%%%%%%% REFERENCES
\newpage
\newpage
{\small
\normalem
\bibliographystyle{ieee_fullname}
\bibliography{egbib}
}

\end{document}